\documentclass[11pt]{article}

\usepackage[english]{babel}

\usepackage[margin=1in]{geometry}
\usepackage{amsmath}
\usepackage{bm}
\usepackage{graphicx}
\usepackage[colorlinks=true, allcolors=blue]{hyperref}
\usepackage{amssymb}
\usepackage{amsthm}
\usepackage{authblk}

\usepackage{booktabs}
\usepackage{multirow}
\usepackage{titlesec}
\usepackage[lined,boxed,linesnumbered,algoruled]{algorithm2e}
\usepackage{cleveref}
\usepackage{algpseudocode}
\usepackage{tikz}
\usetikzlibrary{arrows,shapes,calc,plotmarks,positioning}
\usepackage[lofdepth,lotdepth]{subfig}
\usepackage{dblfloatfix}
\usepackage{lscape}
\usepackage[labelformat=original, labelfont=bf, textfont=normalfont, labelsep=quad, figurename=Fig.]{caption}

\theoremstyle{definition}
\newtheorem{theorem}{Theorem}[section]

\newtheorem{proposition}[theorem]{Proposition}
\newtheorem{definition}[theorem]{Definition}

\newtheorem{example}[theorem]{Example}
\newtheorem*{example*}{Example}
\theoremstyle{remark}
\newtheorem{remark}[theorem]{Remark}

\newcommand{\mr}{\mathbb{R}}

\newcommand{\qaup}{\text{qAP}^\uparrow}
\newcommand{\qadown}{\text{qAP}^\downarrow}
\DeclareMathOperator{\low}{low}
\DeclareMathOperator{\Pert}{Pert}

\titlespacing*{\section}
{0pt}{2.3ex plus 1ex minus .2ex}{1.0ex plus .2ex}
\titlespacing*{\subsection}
{0pt}{2.3ex plus 1ex minus .2ex}{1.0ex plus .2ex}

\titleformat{\section}
  {\normalfont\fontsize{15}{15}\bfseries}{\thesection}{1em}{}

\titleformat{\subsection}
  {\normalfont\fontsize{11}{11}\bfseries}{\thesubsection}{1em}{}

\font\myfont=cmr12 at 16pt
\title{{\myfont Unreduced Persistence Diagrams for Topological Machine Learning}}

\author[1]{Nicole Abreu}
\author[1]{Parker B. Edwards}
\author[1]{Francis C. Motta}
\affil[1]{Department of Mathematics and Statistics, Florida Atlantic University, Boca Raton, FL}
\date{}

\begin{document}

\maketitle

\begin{abstract}
Supervised machine learning pipelines trained on features derived from persistent homology have been experimentally observed to ignore much of the information contained in a persistence diagram. Computing persistence diagrams is often the most computationally demanding step in such a pipeline, however. To explore this dynamic, we introduce several methods to generate topological feature vectors from unreduced boundary matrices and investigate their theoretical and computational properties. We compared the performance of pipelines trained on vectorizations of unreduced PDs to vectorizations of fully-reduced PDs across several data and task types. Our results indicate that models trained on PDs built from unreduced diagrams can perform on par and even outperform those trained on fully-reduced diagrams on some tasks. We also benchmarked the computational performance of an algorithm for computing unreduced diagrams, which was implemented as a heavily modified version of Ripser. These computations are parallelizable and required an order of magnitude less memory on average compared to computing full persistence diagrams. Our results suggest that machine learning pipelines which incorporate topology-based features may benefit in terms of computational cost and performance by utilizing information contained in unreduced boundary matrices. 
\end{abstract}

\section{Introduction}
Topological feature extraction followed by subsequent machine learning (ML) comprises one of the now standard approaches to practical topological data analysis. Successful analyses following this strategy have been conducted on a wide range of data types in diverse scientific domains, including biology~\cite{Amezquita2021Seeds,Benjamin2024,Edwards2021,thomas2021topological}, biomedicine~\cite{chulian2023shape,Nguyen2019}, neuroscience~\cite{Abdallah2023,giustipnas}, finance~\cite{Baitinger2021,GIDEA2018820}, cosmology~\cite{phincosmo,Yip_2024}, material science~\cite{Buchet2018,MOTTA201817}, and many others~\cite{Giunti23a,pmlr-v235-papamarkou24a}. Many of these applications use persistent homology (PH) to quantify topology in their inputs but are limited by practical constraints, particularly due to computational costs. Despite substantial and ongoing improvements since its introduction (e.g., \cite{bauer2021ripser,bauer2022keeping,bauer2017phat,henselmanghristl6,Mischaikow2013Perseus,zhang_et_al:LIPIcs.SoCG.2020.70}), PH remains expensive to compute in many instances~\cite{roadmap}. 

Most combined PH and supervised ML pipelines follow a serial procedure. First, extract topological information from a data set by computing PH for all its entries. The resulting summaries, persistence diagrams (PDs), are then transformed into vectors lying in a Euclidean space. These vectors together with any data labels are then used to train downstream ML models. 

An interesting, if not altogether surprising, phenomenon has been observed about this PH-ML pipeline: much of the information contained in vectorized PDs does not, after training, contribute to ML performance. For example, Bendich et al. found in a regression experiment that vectorizing a PD by retaining the persistence of the 28th most persistent point in their diagrams yielded ``near-optimal'' correlation with their regression target~\cite[Fig. 12]{bendich2016}. In a benchmark of vectorization methods, Ali et al. reported the best performance was obtained from a ``rather na\" ive [vectorization] obtained by collecting basic statistical quantities''~\cite{ali2023survey} from a PD. We report another example of information-discarding behavior in~\Cref{sec:results}.
Because pipelines trained with PH feature vectors often ignore much of the information encoded in a PD, or perform well with vectorizations that discard information before training, an obvious hypothesis is that significant computation is wasted fully computing PH from inputs in these instances. This step often demands the most computational resources in a combined PH-ML pipeline. It is natural to ask whether, rather than discarding information at the vectorization or ML step, one may instead compute approximate or partial PH information without significantly degrading downstream ML performance. If so, can this be done while using fewer resources than a full PH computation? We propose and investigate here a solution which avoids boundary matrix reduction entirely. Our proposed approximation is via \emph{unreduced persistence diagrams}, which are multisets of pairs computed from unreduced boundary matrices. These matrices are the typical starting input that represents a simplex-wise filtration of a complex and, when reduced, produce standard PDs. 

Our approach is predicated on simple observations that recur in several reports on optimized PH algorithms. Namely, in practice, a substantial proportion of columns in PH boundary matrices can be identified with inexpensive computations as encoding persistence pairs or approximate persistence pairs without further reduction. In Bauer's terminology~\cite{bauer2021ripser}, these correspond to \emph{apparent pairs} (APs), which are characterized by the following pattern in a filtration (co)boundary matrix. 
\[
        \begin{array}{c@{}c}
          \begin{array}{c}
               \\
            \\               
             \\
               \\
               \\
               \\
          \end{array}
          &
          \left(
          \begin{array}{ccccc}                
                &        &        &        &       \\
            0   & \cdots &    0   & 1      &      \\
                &        &        & 0      &      \\
                &        &        & \vdots &     \\
                &        &        & 0      &        
          \end{array}
          \right)
        \end{array}
        \]

If a nonzero entry has only 0 entries to its left, it will not be changed during left-to-right matrix reduction. If, additionally, that entry is the lowest nonzero entry in its column, then it corresponds to a persistence pair after reduction. We use both of these conditions independently to define our unreduced PDs. To the best of our knowledge, constructing approximate PDs this way has not been investigated previously.

In \Cref{sec:background} we provide formal definitions for unreduced PDs and collect some of their basic properties. \Cref{sec:stability} investigates their stability both theoretically and experimentally. From the theoretical perspective, most unreduced PD transformations are only locally stable with respect to the bottleneck distance. Only a naive diagram construction consisting of unaltered low 1's from an unaltered boundary matrix is globally stable (\Cref{prop:local_stability} and \Cref{example:instability}). However, for those constructions which fail to be globally stable, we find some evidence that, in practice, stability is maintained under larger perturbations than those under which our results guarantee stability.

\Cref{sec:compute} reports on computational resource benchmarking. We implemented a substantially modified version of Ripser which computes \emph{upper quasi-apparent} PDs (\Cref{def:diagrams}) for Vietoris-Rips (VR) complexes. The algorithm easily parallelizes and generally demands substantially less memory than full reduction via Ripser. The implementation is available at \url{https://github.com/P-Edwards/quasi-apparent-rips}.

Finally, in \Cref{sec:results}, we discuss results from several PH-ML experiments which compare performance when using unreduced PDs against fully reduced diagrams. These experiments were chosen to range in data type and difficulty. The first is a relatively straightforward point cloud shape classification experiment, the second is the standard image classification problem on the Fashion-MNIST~\cite{xiao2017fashionmnist} dataset, and the third is a regression problem using brain scan data processed by Bendich et al.~\cite{bendich2016}. In all experiments, at least one type of unreduced PD performed as well or better than fully reduced diagrams. These results were robust to changes in vectorization method and filtration type. 

\section{Background and Preliminaries}
\label{sec:background}

The theory of PH can be cast in purely algebraic or categorical terms~(e.g.,~\cite{bubenik2014}), but is most easily motivated geometrically. In the latter case, one constructs a finite abstract complex $\Sigma$, cubical or simplicial, from input data. This complex is equipped with a \emph{filtration function} $f:\Sigma\to\mathbb{R}$ which respects face relations from $\Sigma$ in the sense that all faces of an element $\sigma\in\Sigma$ map under $f$ to values at most $f(\sigma)$. Common filtrations include the VR and alpha filtrations for simplicial complexes built on point-cloud data (see~\cite[Chap.~III.2,4]{edelsharer}) and the lower-star filtration built for cubical complexes constructed from image data~\cite{gudhi:CubicalComplex}.

The process of computing a PD usually begins by transforming a filtered complex into a \textit{boundary matrix} $\bm{M} \in \mathbf{Z}_2^{n \times n}$. Call a total ordering on a filtered complex $\Sigma$ \emph{consistent} with a filtration function $f:\Sigma\to\mathbb{R}$ if $f$ is an order-preserving function with respect to the total order on $\Sigma$ and the standard order on $\mathbb{R}$. Arranging the simplices of $\Sigma$ by ascending order subsequently defines an ordered basis for the chain vector spaces $C_*(\Sigma;\mathbf{Z}_2)$. The boundary matrix $\bm{M}$ for a total order consistent with $f$ is the matrix representation of the boundary operator with respect to this ordered basis. We say that a boundary matrix is consistent with $f$ if it is constructed from a total order consistent with $f$.

Having constructed a boundary matrix, one then proceeds to reduce this matrix using a column-wise variant of typical matrix reduction algorithms. We give an overview of the ``standard'' algorithm here, as it motivates our unreduced PD constructions.

First, define $\low(\bm{M}_j) = \max (\{i \; | \; \bm{M}_{i,j} = 1\})$ with $\low(\bm{M}_j)=-1$ if $\bm{M}_j$ is a zero column. Reduce each column of a boundary matrix (proceeding from left to right), by replacing the $j$-th column ${\bm{M}}_j$ with the mod-2 sum $\bm{M}_i + \bm{M}_j$, whenever $\low(\bm{M}_i) = \low(\bm{M}_j)$ and $i<j$, until the low 1 in column $j$ is distinct from all low 1s in the preceding columns or column $j$ consists of all 0 entries~\cite{edelsharer,persalg}. The result is a matrix, ${\bm R}$, of the same size as $\bm{M}$, whose non-zero columns contain distinct low 1s. 

After reduction, the matrix ${\bm R}$ may have some columns consisting entirely of 0s. These are \textit{positive} columns which correspond to the ``birth'' of a homology class. \textit{Negative} columns contain 1s, and correspond to the ``death'' of a homology class. One can prove (e.g., \cite[p.154]{edelsharer}) that the row index $\low(\bm{R}_j)$ for a negative column is the index of a positive column.  

Another important row index associated to columns of a boundary matrix is given by 
\[ 
\beta(\bm{M}_j) := \begin{cases}
                  z, &  \text{if } z \text{ is the largest row index of a 1 } \\ & \text{in column } \bm{M}_j \text{ such that } 
                  \\ & \bm{M}_{z,j'}=0 \text{ for } j'<j, \\ 
                  -1, & \text{if there is no such row index } z
            \end{cases}.
\]
Since reduction of a column is achieved through repeated mod-2 additions with preceding columns, a 1 in row $z$ of column $\bm{M}_j$ cannot be eliminated by reduction if $\beta(\bm{M}_j) = z$. Thus, if $\beta(\bm{M}_j) \neq -1$, column $\bm{R}_j$ is necessarily a negative column \cite{mendoza-smith2017a}. If $\beta(\bm{M}_j) = -1$, column $\bm{R}_j$ may be either negative or positive after reduction. Moreover, since a mod-2 addition of $\bm{M}_j$ with a preceding column can only decrease $\low(\bm{M}_j)$, it follows~\cite[Thm. 5]{mendoza-smith2017a} that $\low(\bm{R}_j) \in [\beta(\bm{M}_j), \low(\bm{M}_j)]$. When $\beta(\bm M_j)\not=-1$, we call the pairs $(\beta(\bm M_j),j)$ and $(\low(\bm M_j),j)$ \emph{upper quasi-apparent} and \emph{lower quasi-apparent} respectively, as $\beta(\bm M_j)$ is the highest possible and $\low(\bm M_j)$ the lowest possible row index for the row which pairs with the negative column $\bm R_j$. 

In the special case where $\beta(\bm{M}_j)= \low(\bm{M}_j)$, one has that $\bm{M}_j = \bm{R}_j$ without any column additions of preceding columns, i.e., $\bm{M}_j$ is fully reduced. The corresponding AP~\cite{bauer2021ripser}, $(\low(\bm M_j),j)$, is thus guaranteed to be a point in the fully reduced PD. This observation has been exploited to great effect in the pursuit of more efficient reduction algorithms~\cite{bauer2021ripser,mendoza-smith2017a,mendoza2017parallel,zhang_et_al:LIPIcs.SoCG.2020.70}. 

Altogether, these observations suggest several possible ways to construct multisets of 
pairs from a boundary matrix $\bm{M}$ which relate to persistence. We study five variants here. 
\begin{definition}
    Let $\bm{M}$ be a boundary matrix consistent with a filtration function $f:\Sigma\to\mathbb{R}$ and let $\bm{R}$ be the result of applying the standard reduction algorithm to $\bm M$. \\
    The \emph{fully reduced persistence diagram} (\text{FR}) of $\bm M$ is \[\text{FR}(\bm{M}) := \{(\low(\bm{R}_j),j)\mid \low(\bm{R}_j)\not= -1\}\cup \{(i,\infty) \mid \low(R_i) = -1, \nexists j(j = \low(\bm{R}_j) \}.\]
    The \emph{apparent pair}  PD   of $\bm M$ is 
    \[ 
    \text{AP}(\bm M) := \{(\low(\bm{M}_j),j) \mid  \beta(\bm{M}_j) = \low(\bm{M}_j) \not= -1 \}.
    \]
    The \emph{lower quasi-apparent pair} (\text{$\qadown$}) PD of $\bm M$ is
    \[ 
    \qadown(\bm M) := \{(\low(\bm{M}_j),j) \mid \beta(\bm{M}_j) \not= -1 \}.
    \]
    The \emph{upper quasi-apparent pair} (\text{$\qaup$}) PD of $\bm M$ is
    \[ 
    \qaup(\bm M) := \{(\beta(\bm{M}_j),j) \mid  \beta(\bm{M}_j) \not= -1 \}.
    \]
    The \emph{low-ones} (\text{L1}) PD of $\bm M$ is 
    \[ 
    \text{L1}(\bm M) := \{(\low(\bm{M}_j),j) \mid \low(\bm{M}_j) \not= -1 \}.
    \]
    The latter four PDs are \emph{unreduced} persistence diagrams. \label{def:diagrams}
\end{definition}

Note that each of these PDs may be translated from indexed-based PDs to filtration-value PDs using filtration values for the columns. That is to say, an indexed-based PD consisting of index pairs $(i,j)$ is mapped via a filtration function to the multiset with pairs $(f(\sigma_i),f(\sigma_j))$ when $j < \infty$ and $(f(\sigma_i),\infty)$ when $j=\infty$. Each point $(i,j)$ in a PD has an associated homology dimension $\dim(\sigma_i)$. We will occasionally, by an abuse of terminology, use the same abbreviations in \Cref{def:diagrams} to refer to a PD after the filtration function has been applied. In that case, we do not include the \emph{ephemeral} pairs where $f(\sigma_i) = f(\sigma_j)$ in any diagram. This is because ephemeral pairs correspond to 0-\textit{persistence} ($f(\sigma_j)-f(\sigma_i) = 0$) PH classes, and so typically do not contribute information to a vectorization of a PD.

Ephemeral pairs arise naturally in both fully reduced and unreduced PDs. For example, in the VR construction, the filtration function $f:\Sigma\to\mathbb{R}$ built from point cloud data is defined on simplices of dimension higher than 1 as the largest filtration value of a simplex's 1-dimensional faces, i.e., $f(\sigma) := \max\{f(\sigma') \mid \sigma'\subseteq\sigma, \; \dim(\sigma') =1 \}$. After sorting by filtration value, it follows that if $\bm{M}_j$ is a column corresponding to a simplex $\sigma_j$ with dimension greater than 1, $f(\sigma_i) = f(\sigma_j)$ when $i = \low(\bm{M}_j)$. The filtration-mapped low-ones diagram for $\bm M$ is therefore empty in homology degrees greater than 0. This extends to all other unreduced diagrams except $\qaup(\bm M)$ in light of the following simple containments. 

\begin{proposition}
\label{prop:pd_containments}
    Let $\bm M$ be a boundary matrix consistent with a filtration function $f:\Sigma\to\mathbb{R}$ and let $\bm M^{(k)}$ be the matrix obtained from $\bm M$ after reducing the first $k$ columns via the standard reduction algorithm. Then the following inclusions of multisets hold
    \[ 
    \text{AP}(\bm M^{(k)}) \subseteq \qadown (\bm M^{(k)}) \subseteq \text{L1}(\bm M^{(k)})
    \]
    \[ 
    \text{AP}(\bm M^{(k)}) \subseteq \qaup (\bm M^{(k)}) 
    \]
    \[ 
    \text{AP}(\bm M^{(k)}) \subseteq FR(\bm M).
    \]
\end{proposition}

\begin{remark}
    Reprising the notation of \Cref{prop:pd_containments} with $\bm R$ the reduced boundary matrix, and denoting by $T$ any unreduced PD transformation, one has directly that $d_B(T(\bm M^{(k)}), T(\bm R))$, where $d_B$ denotes the bottleneck distance, decreases to $0$ monotonically as $k$ increases. Observe, however, that if one excludes points with $\infty$ as a coordinate that $\text{L1}(\bm R) = \text{FR}(\bm M)$. This is not the case in general for any other unreduced PD transformation.
\end{remark}

One of the main goals of our unreduced PD constructions is to save resources when computing them, particularly memory, compared to full reduction. There are, broadly speaking, three primary objects which, when stored, drive memory usage for implementations of the standard algorithm: the filtration boundary matrix, the current column being reduced as previous columns are added to it, and a set of already fully reduced columns or low 1 indices for such columns. In many important instances, e.g., when using VR complexes~\cite{bauer2021ripser,zhang_et_al:LIPIcs.SoCG.2020.70}, one may avoid saving an explicit representation of the filtration boundary matrix as well as the low 1 indices for a large proportion of reduced columns. As they do not require reduction, unreduced diagrams additionally avoid storing working (co)-boundaries throughout reduction for columns. We study the differences this makes in practice with an implementation of an algorithm for computing $\qaup$ diagrams of VR complexes in \Cref{sec:compute}.

\section{Stability} \label{sec:stability}
It is natural to ask whether our unreduced PD transformations are stable. Stability is, more precisely, a statement about Lipschitz continuity of these transformations. A PD transformation $T$ is \textit{stable} if there is $C>0$ such that $d_B(T(M^f),T(M^g)) \leq C\Vert f-g\Vert_\infty$ for any two filtration functions $f,g:\Sigma\to\mathbb{R}$ on the same complex $\Sigma$ and any two boundary matrices $M^f,M^g$ consistent with $f$ and $g$ respectively. Here $T$ indicates transforming via a reduced or unreduced PD construction. Unless otherwise indicated, we consider all homology degrees simultaneously. 

The extended pseudo-metric $d_B$ is the \emph{bottleneck distance}. A celebrated and now foundational result in PH theory is that the fully reduced diagram transformation is stable with Lipschitz constant $C=1$~\cite{bauerlesnickstability2015,chazalstructurestability2016,csehstability2007}. We will see that the situation for unreduced diagrams is more subtle. 

\subsection{Theoretical results}
Recall the definition of the bottleneck distance. First, a \emph{partial matching} between two finite multisets of pairs $D_1,D_2$ is an injective map $\varphi:S\to D_2$ where $S$ is a subset of $D_1$. The cost of the partial matching $\varphi$, $\text{cost}_\infty(\varphi)$, is defined as the maximum value in the three sets $\{\Vert p - \varphi(p)\Vert_\infty \mid p\in S\}$, $\{ \Vert p-\Delta(p) \Vert_\infty \mid p\in D_1-S\}$, and $\{\Vert p'-\Delta(p')\Vert_\infty \mid p'\in D_2-\text{im}(\varphi)\}$. For any point $(b,d)\in\mathbb{R}^2$, $\Delta((b,d)):= (\frac{b+d}{2},\frac{b+d}{2})$ is the closest point to $(b,d)$ on the diagonal. We sometimes say that the points in $S$ and $\text{im}(S)$ are matched, whereas the points which are not in these sets are matched to the diagonal. Finally, define
\[ 
d_B(D_1,D_2) := \min \{ \text{cost}_\infty(\varphi) \mid \varphi \text{ is a partial matching } D_1\to D_2\}.
\]
When considering PDs $D_1, D_2$ with points of multiple dimensions $1,\dots,d$, define $d_B(D_1,D_2) := \max_{k=1,\dots,d} d_B(D_1^k,D_2^k)$ where $D_*^k$ denotes the subset of the diagram consisting of points with dimension $k$. 

If a filtration function $f:\Sigma\to\mathbb{R}$ is not injective, then there is not a unique boundary matrix consistent with $f$. This causes no issue when discussing fully reduced PDs, as standard results guarantee that there is a unique PD obtained from reducing any boundary matrix that is consistent with $f$. A similar result does not hold for unreduced PDs generally, so we will discuss stability for these constructions relative to boundary matrices. 

In the case of a low-ones diagram for a boundary matrix, there is a bijection between points in the diagram and columns of the matrix. When the boundary matrix is constructed from a filtered complex, the columns are also in bijection with the simplices of the complex. We can compose the bijections to obtain a bijection between low-ones diagrams of different filter functions on the same complex, which is enough for stability.
\begin{proposition}\label{prop:l1_stability}
    Let $\Sigma$ be a finite complex and $f,g:\Sigma\to\mathbb{R}$ be any two filtration functions. If $M^f$ and $M^g$ are any two boundary matrices consistent with $f$ and $g$ respectively and $L1(M^f)$, $L1(M^g)$ are their low-ones diagrams, then $d_B(L1(M^f),L1(M^g)) \leq \Vert f-g\Vert_\infty$.
\end{proposition}
\begin{proof}
   For $\sigma\in\Sigma$, let $M_\sigma^f,M_\sigma^g$ denote the columns corresponding to $\sigma$, and for a row or column index $i$ of a boundary matrix let $\Sigma_i$ denote the corresponding simplex. Define $\tau_f = \Sigma_{\low(M^f_\sigma)}$ and define $\tau_g = \Sigma_{\low(M^g_\sigma)}$ similarly. Consider the partial matching $\varphi:L1(f)\to L1(g)$ defined by $(f(\tau_f),f(\sigma))\mapsto (g(\tau_g),g(\sigma))$, where $(f(\tau_f),f(\sigma))$ is unmatched if $(g(\tau_g),g(\sigma))$ is ephemeral. Setting $\epsilon = \Vert f-g\Vert_\infty$, we claim that $\text{cost}_\infty(\varphi)\leq \epsilon$. It follows by definition that $\vert f(\sigma) - g(\sigma)\vert \leq \epsilon$. Without loss of generality assume that $g(\tau_g) \leq f(\tau_f)$. By definition of $\low$, $g(\tau_f)\leq g(\tau_g)$ and so also $g(\tau_f) \leq f(\tau_f)$. Subsequently, $0 \leq  f(\tau_f) - g(\tau_f) \leq \epsilon$ and thus $[f(\tau_f) - g(\tau_g)] + [g(\tau_g)-g(\tau_f)] = f(\tau_f)-g(\tau_f) \leq \epsilon$. Since $g(\tau_g)-g(\tau_f) \geq 0$, we have $f(\tau_f) - g(\tau_g) \leq \epsilon$. 
\end{proof}

Global stability no longer holds for quasi-apparent and apparent pair diagrams. As the next example shows, the main obstruction to this is combinatorial. Small changes in the filtration function can reorder the rows and columns of an induced boundary matrix in a way that deletes apparent and quasi-apparent columns. 

\begin{example}
 Consider the following set of points in $\mr$
    \begin{center}
        $P_\gamma = \{ 0,2,5+\gamma,8 \}$
    \end{center}
    where $\vert\gamma\vert \leq 3$. Labeling the points $A,B,C,D$ in the order they appear above, the distance matrix for the point cloud is
    \[
    \bordermatrix{
      & A & B & C & D \cr
    A & 0 & 2 & 5+\gamma & 8 \cr
    B &   & 0 & 3+\gamma & 6 \cr
    C &   &   & 0        & 3-\gamma \cr
    D &   &   &          & 0 \cr
    }
    \]
    
    Let $\Sigma$ be the complete 1-dimensional simplicial complex on vertex set $\{A,B,C,D\}$. Define $f_\gamma(v) = 0$ for every vertex $v$ and $f_\gamma(uv) = \vert u - v\vert$ for any 1-simplex $uv$. A boundary matrix consistent with $f_\gamma$ is depicted on the left of row (b) in \Cref{fig:Instability_Example} and a boundary matrix consistent with $f_{-\gamma}$ is depicted on the right. The corresponding AP, $\qaup$ and $\qadown$ diagrams for $f_\gamma$ are
    \[ 
    \text{AP}_\gamma = \qadown_\gamma  = \qaup_\gamma = \{(0,2), (0,3-\gamma)\}.
    \] 
    Similarly for $-\gamma$, we have
    \[\text{AP}_{-\gamma} = \qadown_{-\gamma}  = \qaup_{-\gamma} = \{(0,2), (0,3-\gamma),(0,3+\gamma)\}.\] The additional point that appears due to a column order switch between 1-simplices $BC$ and $CD$ is highlighted with a red box in Figure~\ref{fig:Instability_Example} in the two right-most PDs in rows (b) and (c). For all sufficiently small $\gamma > 0$, we have that $\Vert f_\gamma - f_{-\gamma}\Vert_\infty = 2\gamma$ while $d_B(D_\gamma,D_{-\gamma}) = \frac{3+\gamma}{2} > \frac{3}{2}$. Here $D$ indicates any of the unreduced diagrams $\text{AP},\qadown,\qaup$. 
    
    \begin{figure}[t]
        \centering
        \includegraphics[width=0.9\linewidth]{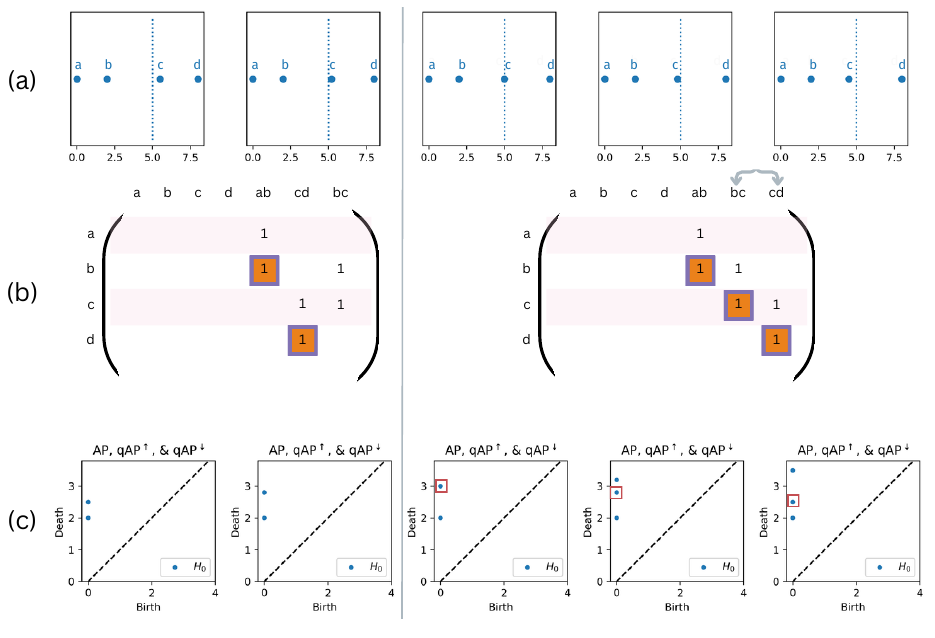}
        \caption{Example of instability for the $\qadown$, $\qaup$, and AP diagrams. A point cloud (a) of 4 points is generated and, from left to right, point $c$ is moved closer to point $b$. $\qadown$, $\qaup$, and AP diagrams (each unreduced PD type is exactly the same in this example) of each point cloud is displayed directly below the corresponding point cloud in row (c). Row (b) illustrates the column swap caused by the movement of point $c$ in row (a). The additional point in each unreduced PD caused by the column swap illustrated in row (b) is highlighted by a red box}
        \label{fig:Instability_Example}
    \end{figure}
    \label{example:instability}
    
\end{example}

At the very least, the behavior in these examples cannot occur without reordering of the rows and columns. We can use this observation directly to show that the QA and AP transformations are locally Lipschitz. 

\begin{proposition}
    Let $f,g:\Sigma\to\mathbb{R}$ be filtration functions on a finite complex $\Sigma$ with $f$ injective. Enumerate the simplices of $\Sigma$ in filtration order by $f$ as $\sigma_1,\sigma_2,\dots,\sigma_N$. Take $\epsilon$ to be half the minimum gap between adjacent simplices, $\epsilon := \min_{i=1}^{N-1} [f(\sigma_{i+1}) - f(\sigma_i)]/2$. If $\Vert f - g \Vert_\infty < \epsilon$, then $g$ is also injective and $d_B(T(M^f),T(M^g)) \leq \Vert f - g\Vert_\infty$ where $T\in\{\qaup, \qadown, \text{AP}\}$ and $M^f, M^g$ denote the uniquely determined boundary matrices corresponding to $f$ and $g$.\label{prop:local_stability}
\end{proposition}
\begin{proof}
   Suppose $\Vert f - g \Vert_\infty<\epsilon$. Note that since $f$ is injective, there is a unique boundary matrix $M^f$ associated to $f$. Similarly, we claim that $g$ is injective. If, to the contrary, $g(\sigma_k) = g(\sigma_j)$ for some $\sigma_k\not=\sigma_j$ with $k > j$, then $g(\sigma_k) = g(\sigma_{k-1})$ and so $\vert f(\sigma_k)-f(\sigma_{k-1})\vert \leq \epsilon$ using the triangle inequality. This contradicts the definition of $\epsilon$. Therefore there exists a unique boundary matrix $M^g$ associated to $g$. Notice that the total orderings on $\Sigma$ induced by $f$ and $g$ are the same, so $M^f = M^g$. We obtain a bijection between $T(M^f)$ and $T(M^g)$ by matching the point in $T(M^f)$ for a negative column in $M^f$ with the point in $T(M^g)$ for the column in $M^g$ with the same index. This matching has cost at most $\Vert f - g\Vert_\infty$. 
\end{proof}

\begin{remark}
    For a fixed finite complex $\Sigma$, we may identify the set of filtration functions on $\Sigma$ with a semialgebraic subset of $\mathbb{R}^{\vert \Sigma \vert}$ with positive measure. The set of non-injective functions are identified with tuples which share the same value for at least two coordinates. This is a proper algebraic subset and so is measure 0. Thus the set of non-injective functions is measure 0, and the quasi-apparent and apparent pair diagrams constructions are locally Lipschitz generically, i.e., for any filtration function outside of a set of measure 0. 
\end{remark}

\subsection{Experimental results}
The local stability radius $\epsilon$ in \Cref{prop:local_stability} is pessimistic in the sense that we took $\epsilon$ small enough so that the boundary matrix for any filtration in a ball of radius $\epsilon$ around a filtration function $f$ is the same as the boundary matrix for $f$. Focusing on the instabilities exhibited in \Cref{example:instability}, however, notice that row and column ordering swaps do not always delete apparent or quasi-apparent pairs. Furthermore, even if swapping the order of two simplices by perturbing $f$ deletes an AP or QA pair and leaves the other pairs intact, the deleted pair may have persistence too low to cause an unstable change, or any change at all, in the bottleneck distance. 

We therefore performed a computational experiment to explore the stability behavior of AP/QA diagrams in more detail. The FR Rips PD transformation is stable with respect to the bottleneck distance, with Lipschitz constant $C = 2$ relative to the Gromov-Hausdorff distance~\cite{chazal2013persistencestabilitygeometriccomplexes}. In our experiments, all point clouds lay in the same ambient Euclidean metric space, allowing us to use the more computable Hausdorff distance,
\[ d_H(P_i,P_j) = \max \left\{ \sup_{p_i\in P_I} \inf_{p_j\in P_j} \|p_i - p_j\|_\infty, \sup_{p_j\in P_j} \inf_{p_i\in P_i} \|p_j - p_i\|_\infty \right\}, \]
where $p_i \in P_i, p_j \in P_j$, and we can expect the same Lipschitz constant to be valid relative to the Hausdorff distance, since $d_{GH}(X,Y) \le d_H(X,Y)$. 
We use the Hausdorff distance to estimate the Lipschitz constant of the Rips AP, $\qadown$, and L1  PDs computationally with the ratio
\[ \frac{d_B(T(\mathbb{R}\text{ips}(P_i)),T(\mathbb{R}\text{ips}(P_j)))}{d_H(P_i,P_j)}. \]
 Large observed values or spikes in observed ratios from a computational experiment correspond to instabilities in the diagram transformation $T$. 

\begin{figure}[t]
    \centering
    \includegraphics[width=0.9\linewidth]{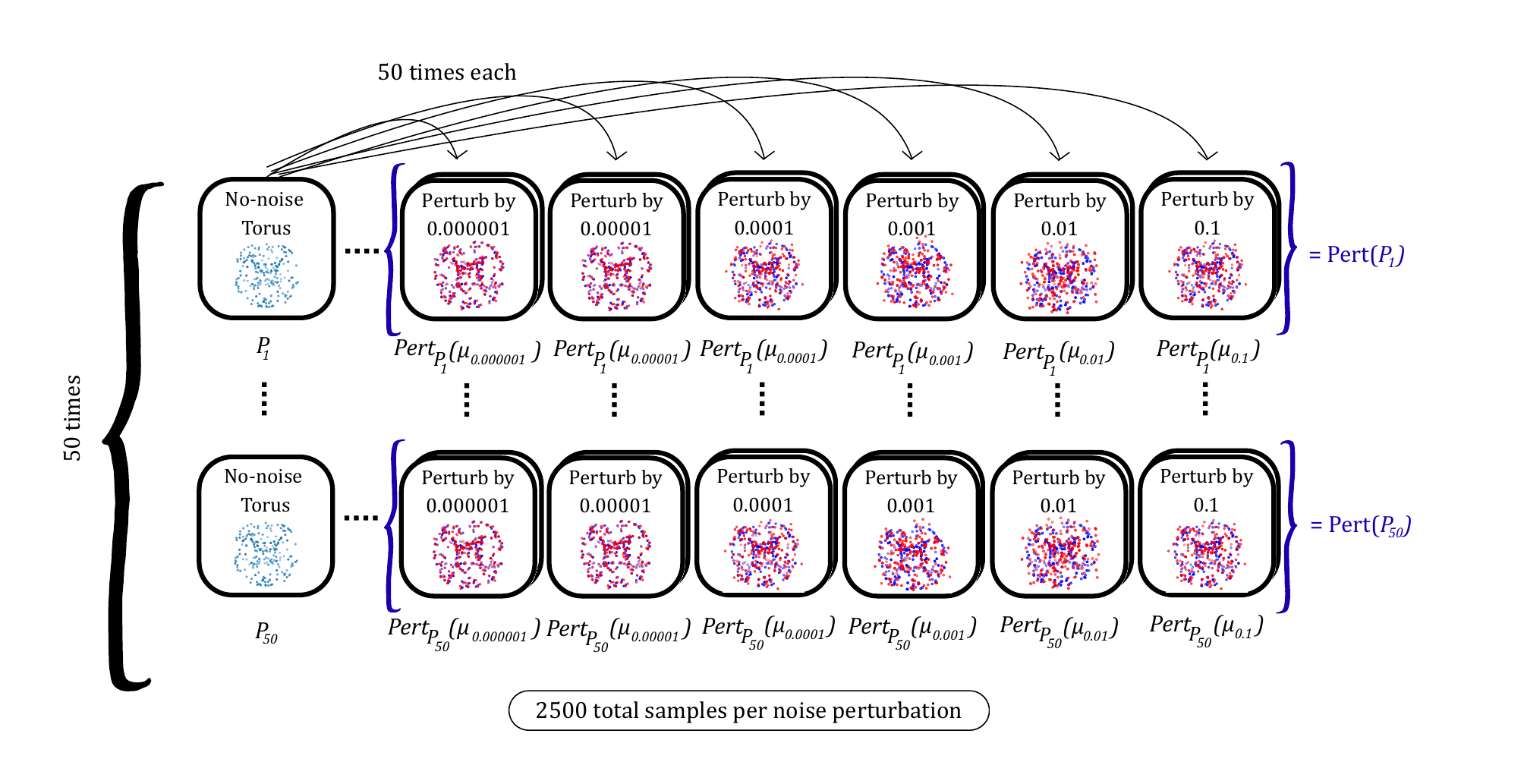}
    \caption{Sampling procedure for the stability experiment }
    \label{fig:Stability_Experiment_PointCloud_Sampling}
\end{figure}

Our experiment proceeded as follows. We first computed a point cloud sample $P$ of 200 points on a unit torus in $\mathbb{R}^3$. Then, for a fixed noise level, $\mu$, we repeatedly created perturbed point cloud samples $S_1,\dots,S_{50}$ of $P$ by perturbing each point in $P$ in a uniform random direction, drawn from the 2-sphere, with a uniform random magnitude drawn from (0,$\mu$). We repeated this for various noise levels $\mu_1, ..., \mu_n$ and constructed the set $ \Pert(P) = \{ \Pert_P(\mu_1), \dots, \Pert_P(\mu_n) \}$ where $\Pert_P(\mu_j) = \{ S_{P,\mu_j,1}, \dots, S_{P,\mu_j,50} \} $ and $S_{P,\mu_j,i}$ is the $i^{\text{th}}$ perturbed point cloud sample of $P$ with noise level $\mu_j$. We then repeated this entire process for 50 points cloud samples, $P_i$, each consisting of 200 points on the unit torus in $\mathbb{R}^3$, to  obtain the set $\{ \Pert(P_1), \dots, \Pert(P_{50}) \}$. An illustration of this process is shown in \Cref{fig:Stability_Experiment_PointCloud_Sampling}.

We then created FR, AP, $\qadown$, and L1 Rips PDs of each of the point clouds in $\Pert(P)$. Finally, we computed the Hausdorff distance between each base point cloud $\Pert(P_i)$ and its perturbations over different noise levels $S_{P_i, \mu_j,k}$ for $1 \leq j \leq n$ and $1 \leq k \leq 50$; computed the bottleneck distances between the PDs of each base point cloud and the PDs of its perturbations over different noise levels; and then recorded the corresponding bottleneck/Hausdorff distance ratios. 

\begin{figure}
    \centering
    \includegraphics[width=\linewidth]{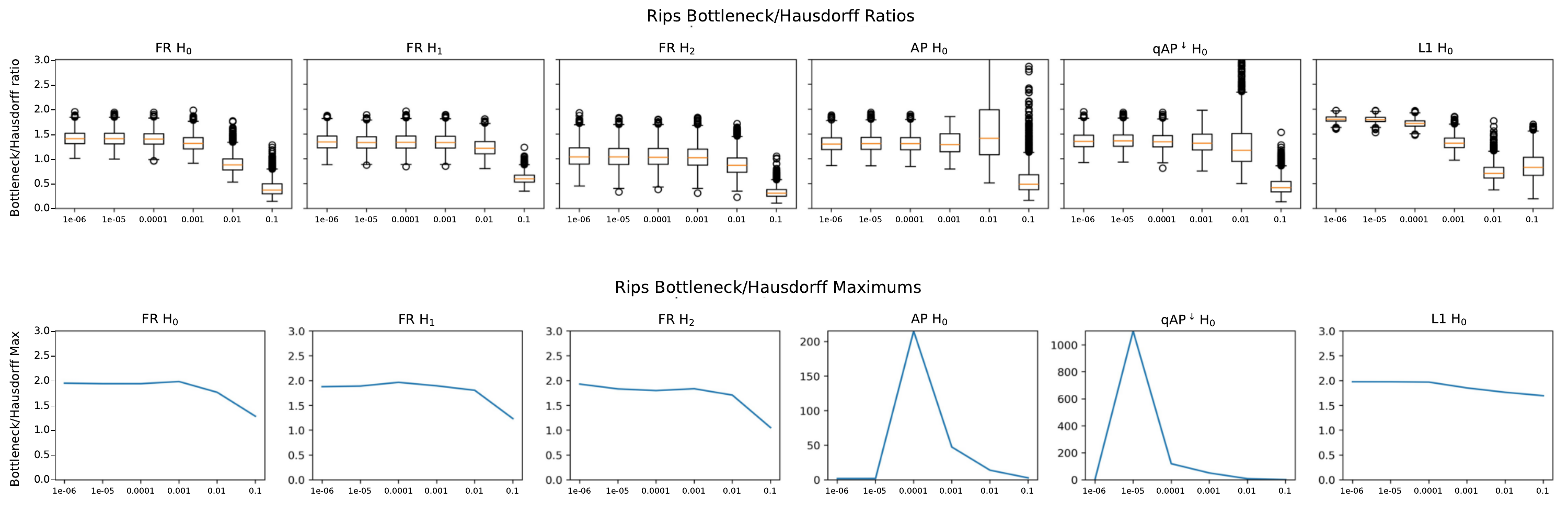}
    \caption{The box-plots (row 1) and the maximum values (row 2) of the bottleneck/Hausdorff distance ratios per noise level. There are 2500 ratios per noise levels. Outliers which exceeded the ratio value 3 were excluded from the plots in row 1}
    \label{fig:Stability_Experiment_Plots}
\end{figure}

For noise levels $\mu_1 = 0.000001, \mu_2 = 0.00001, \mu_3 = 0.0001, \mu_4 = 0.001, \mu_5 = 0.01,$ and $\mu_6 = 0.1$, 50 base samples, and 50 perturbations per noise level of each base sample, we obtained 2500 bottleneck/Hausdorff ratios per noise level. The results are displayed in Figure~\ref{fig:Stability_Experiment_Plots} and show instabilities in the AP PDs for noise levels $\mu = 0.0001, 0.001, 0.01,$ and 0.1 and instabilities in the $\qadown$ PDs for noise levels $\mu = 0.00001, 0.0001, 0.001,$ and 0.01. 

Empirical study confirms theoretical stability results for FR and L1 PDs. Very large ratios of bottleneck to Hausdorff distances can be observed for the AP and $\qadown$ diagrams, but these events appear exceedingly rare, with the majority of the distribution across random perturbations realizing ratios below the FR Lipschitz constant. For example, the top whiskers of the boxplots for both AP and $\qadown$ diagrams at noise level $\mu=0.0001$ in \Cref{fig:Stability_Experiment_Plots} are below 2. 


We also ran an additional series of experiments using the same framework, resampling both point clouds and perturbations at the same noise levels as before only until a ratio value larger than 3 was observed. This was done both for AP diagrams and $\qadown$ diagrams. In all cases, this first observed large ratio value was produced by a diagram which had a different number of points than the unreduced PD of the corresponding unperturbed point cloud. This behavior is consistent with the type of instability discussed in \Cref{example:instability}.

\section{Computational benchmarking} \label{sec:compute}
A main motivation behind investigating unreduced PDs is the potential to save computational resources compared to full PDs. As a proof of concept, we implemented and benchmarked a substantially modified version of Bauer's software Ripser~\cite{bauer2021ripser} to compute $\qaup$ diagrams for VR filtrations of point cloud data. We focus on $\qaup$ diagrams since, as discussed in \Cref{sec:background}, other unreduced PDs for VR complexes have only ephemeral pairs in homology dimensions greater than 0. That implementation is available at {https://github.com/P-Edwards/quasi-apparent-rips}. 

For the sake of comparison with implementations of full reduction algorithms, it is best to describe our approach in terms of the filtration coboundary matrix associated to a simplex-wise filtration (see, e.g., \cite[\S 3.3]{bauer2021ripser} and \cite{des2011dualities}). If $\bm{M}$ is the boundary matrix for homology in degree $d$ associated to a simplex-wise filtration, the filtration coboundary matrix $\bm{M}^*$ is the anti-transpose obtained by transposing $\bm{M}$ and then reversing the order of the rows and columns. Computing the $\qaup$ diagram can be described as a search across the columns of $\bm{M}^*$. We abuse notation slightly and identify a column or row in $\bm{M}^*$ corresponding to a simplex $\sigma$ with $\sigma$. Also, $\low(\cdot,\bm M^*)$ denotes the low function on columns of $\bm M^*$. The following is immediate from the definitions. 

\begin{proposition}
    Let $\bm M$ be the filtration boundary matrix for a simplex-wise filtration, $\bm M^*$ the corresponding filtration coboundary matrix, $\sigma$ a $(d+1)$-simplex, and $\tau$ a $d$-simplex. Then $\beta(\sigma) = \tau$ if and only if $\low(\tau,\bm M^*) = \sigma$. 
\end{proposition}

Our computational approach is summarized in \Cref{alg:qap}. Technical complications introduced by resource-saving strategies unique to VR complexes are omitted. Unrelated to VR-specific strategies, our algorithm stores and updates a representation of the input's $\qaup$ diagram as it progresses. We use a hash map data structure to store this representation. This is naturally suited to the algorithm's structure, which is a search across the columns of a filtration coboundary matrix.

\begin{algorithm}[h!]
    \scriptsize
	\SetKwInOut{input}{Input}\SetKwInOut{output}{Output}\SetKwFunction{Return}{Return}	
    \input{A filtration coboundary matrix $\bm{M}^*$ for cohomology dimension $d$ }
    \input{A filtration function $f$}
    \output{The set of non-ephemeral pairs in the corresponding $\qaup$ diagram.} 
    \smallskip
    current\_pairs = hash map from $(d+1)$-simplices to $d$-simplices \;
    \For{each column of $\bm{M}^*$ with corresponding to $d$-simplex $\tau$}{
        $\sigma = \low(\tau,\bm{M}^*)$ \;
        \If{$\sigma$ is not in current\_pairs or ($\tau' =$ current\_pairs$[\sigma]$ and $f(\tau) > f(\tau')$) }{
            current\_pairs$[\sigma] = \tau$\;
        }

    }
    \Return $\{ (f(\tau), f(\sigma)) \mid$ current\_pairs$[\sigma]=\tau$ and $f(\tau)\not=f(\sigma)\}$
	\caption{\textsc{Compute upper quasi-apparent pair diagram}\label{alg:qap}}	
\end{algorithm}

This algorithm straightforwardly parallelizes\footnote{Our parallel search scheme bears some resemblance to the parallel search for APs implemented by Zhang et al. in Ripser++~\cite{zhang_et_al:LIPIcs.SoCG.2020.70}. Their algorithm stores all APs, however, which requires more memory than the recomputation strategy from Ripser that our implementation adapts.}, as computations from different columns do not affect one another except possibly by modifying entries which share keys in the hash map. The simplest way to avoid race conditions is to maintain a separate copy of the hash map for every computation thread and merge those maps at the end to obtain a final result. Parallelizing a loop over the columns of the coboundary matrix contrasts with the serial standard algorithm. In a nutshell, for full reduction one must instead conduct a similar for loop left-to-right over columns of $\bm{M}^*$. 

For VR complexes, the columns of $\bm{M}^*$ can be enumerated independently of one another by, e.g., assigning each vertex $v$ to a computation thread, processing its column in the loop, then recursively processing the cofaces of $v$ which are in the complex, their cofaces, etc. up to a maximum dimension. One can do this enumeration with minimal memory overhead and without repeating columns~\cite[p. 414]{bauer2021ripser}. Thus the for loop in \Cref{alg:qap} can be executed efficiently in parallel. 

We can also save substantial memory and compute resources by taking advantage of \emph{zero} APs, i.e., ephemeral APs, similar to Ripser. If column $\tau$ in $\bm{M}^*$ is in an ephemeral AP with $\sigma = \low(\tau,\bm{M}^*)$, then $(f(\tau),f(\sigma))$ will not contribute to the final unreduced diagram, so need not be stored. Furthermore, if we have computed $\sigma = \low(\tau,\bm{M}^*)$ and subsequently identify that $\sigma$ is in a zero AP, $(\tau',\sigma)$, then necessarily $f(\tau')\geq f(\tau)$, so we will not store $(\tau,\sigma)$ in the final hash map for \Cref{alg:qap}. Since searching for zero APs is much faster in general for VR complexes than computing $\low(\tau,\bm M^*)$ and many columns are in a zero AP~\cite{bauer2021ripser,zhang_et_al:LIPIcs.SoCG.2020.70}, identifying these circumstances and skipping storage and computation appropriately yields substantial savings.

There is a second type of zero AP column, namely when a $d$-simplex column $\tau$ is in an AP $(\phi,\tau)$ where $\phi$ is a $(d-1)$-simplex. The column for $\tau$ requires no additional processing in the full reduction algorithm as it will be a 0 column in the fully reduced filtration coboundary matrix. The pair $(f(\tau),f(\low(\tau,\bm{M}^*))$ can still by definition appear in the unreduced $\qaup$ diagram, however. We skip columns $\tau$ with this property in our implementation, as Ripser does, thereby losing some information. Performance results using $\qaup$ diagrams with VR complexes for ML experiments in \Cref{sec:results} are reported using diagrams with this choice.

\begin{figure}[t]
    \centering
    \includegraphics[width=0.75\linewidth]{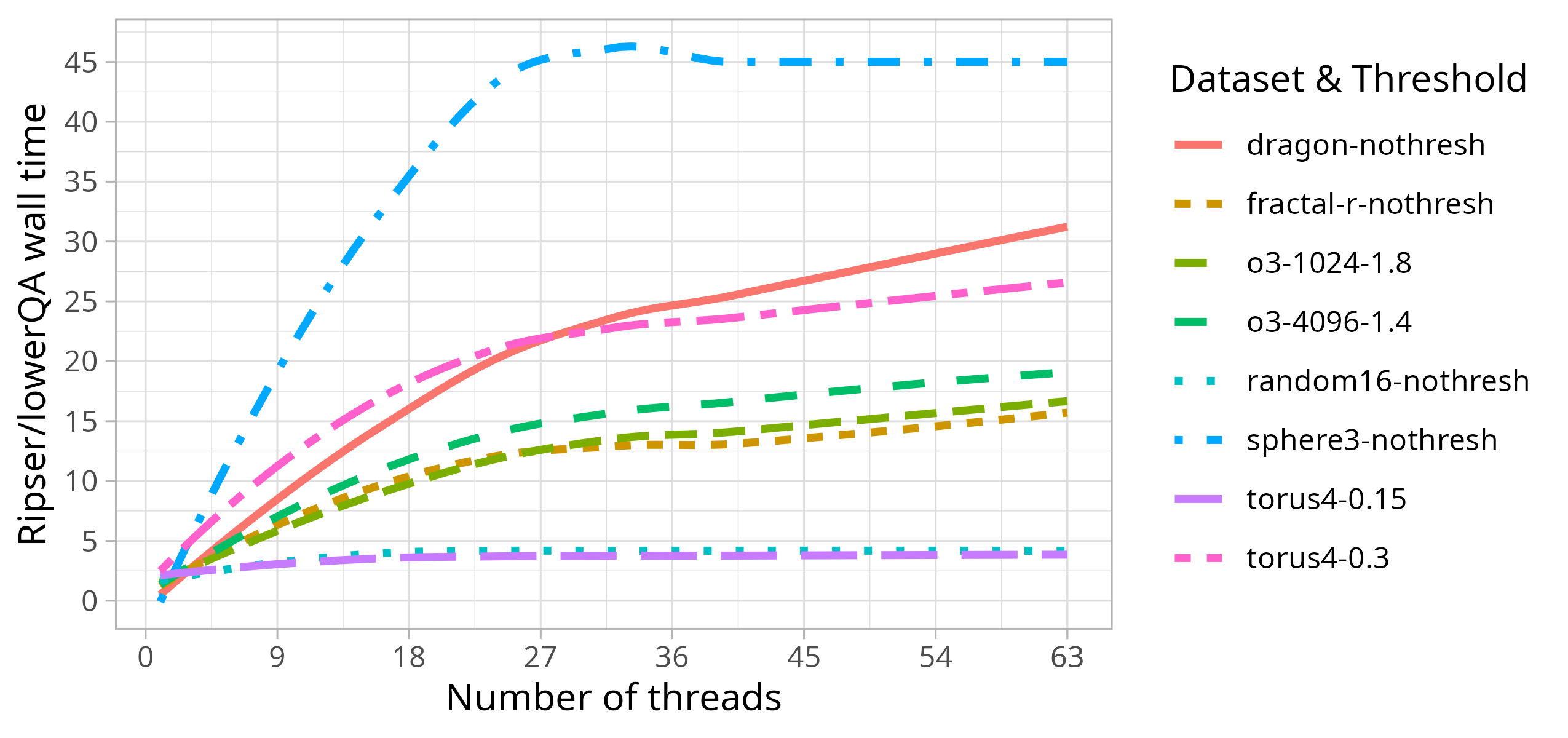}
    \caption{Benchmarking results for wall time comparison for $\qaup$ diagrams vs Ripser. The line plots depicted have been smoothed from the original data to improve legibility. An unsmoothed version of the plot is available in the Appendix as \Cref{fig:unsmoothed_wall_time}}
    \label{fig:walltime_ratios}
\end{figure}

We benchmarked our implementation and Ripser on a computer with a 32-core AMD Ryzen Threadripper PRO 5975WX CPU, 504GB of RAM, and 64 computation threads using several point cloud data sets that have been previously used for this purpose\footnote{We obtained the cyclo-octane configuration space dataset from the copy hosted with Stolz et al.'s work~\cite{stolz2020detection}. It originated from \texttt{JavaPlex}~\cite{adams2014javaplex} and ultimately from a sub-sample of about 1 million configurations originally produced by Martin et al. as reported in their analysis of the cyclo-octane energy landscape~\cite{martin2010topology}. Neither this original data file nor a means to recompute it independently are publicly available to our knowledge.}~\cite{adams2014javaplex,bauer2021ripser,bauer2017phat,roadmap}. Their specifications are collected in \Cref{tab:data_sets}. Since our implementation is multi-threaded, we computed $\qaup$ diagrams for each data set 63 times, once for each possible number of threads available to us (saving one thread for user interaction). Resource consumption was tracked with \texttt{/usr/bin/time}. Results are collected in \Cref{tab:benchmark_results} and \Cref{fig:walltime_ratios}. 

Wall time speed-ups from parallelization were substantial, but with diminishing nonlinear returns\footnote{Some of this nonlinearity occurring over 32 compute threads appears to be due to simultaneous multithreading dynamics, as our testing machine had 32 physical cores. Savings were also not wholly linear from 1 to 32 threads, however.}. Using the 63 threads available to us resulted in on average 20 times less wall time than a full persistence computation using Ripser. This speed-up varied by example, with examples having more simplices generally exhibiting larger speed-ups. Our implementation required comparable or less user time to Ripser in all instances. These results support the hypothesis that $\qaup$ diagrams require less compute time in practice, including for highly-optimized VR persistence computations.

Memory savings were also significant. There are two classes of examples to compare. Both implementations allow a user to specify a radius threshold. In this case, a diagram is computed on the subcomplex of the full VR complex having edges with length at most the threshold. When thresholding this way, both implementations store the 1-skeleton of the VR complex in memory in order to speed up $\low(\cdot,\bm{M}^*)$ computations. If no threshold is provided, diagrams are computed with threshold the minimum enclosing radius of the point cloud, but without storing a 1-skeleton in memory.

Our algorithm used on average 1/13th of Ripser's memory footprint on non-thresholded examples and 1/1.86 on thresholded examples. One should expect this discrepancy when thresholding. Ripser's memory usage typically consists overwhelmingly of 1-skeleton storage on smaller datasets with thresholds.

We also tested two more demanding examples, computing $\qaup$ diagrams up to homology degree 2 and threshold the minimum enclosing radius for the torus4 and cyclo-octane datasets, which contain 50,000 points and 6040 points respectively. The former was too large for Ripser to complete due to memory requirements, while we were able to compute an unreduced diagram in parallel in about 112 hours, i.e., 4 and 2/3rds days. Ripser completed computation on the cyclo-octane dataset, for which a $\qaup$ diagram required 59.5 times less wall time with 63 threads and 454 times less memory than Ripser.

\begin{landscape}
\begin{table}
    \centering
    \begin{tabular}{|cccc|}
            \hline
          & Dimension & Number of points & Max homology degree \\
         torus4~\cite{bauer2021ripser} & 4 & 50,000 & 2 \\
         \hline
         fractal-r~\cite{roadmap} & 259 & 512 & 2 \\
         \hline
         o3-1024~\cite{bauer2021ripser} & 9 & 1024 & 3 \\
         \hline
         o3-4096~\cite{bauer2021ripser} & 9 & 4096 & 3 \\
         \hline
         random16~\cite{roadmap} & 16 & 50 & 7 \\
         \hline
         sphere3~\cite{bauer2017phat} & 3 & 192 & 2 \\
         \hline
         dragon~\cite{roadmap} & 3 & 2000 & 2 \\
         \hline
         cyclo~\cite{adams2014javaplex,stolz2020detection,martin2010topology} & 24 & 6040 & 2 \\
         \hline
    \end{tabular}
    \caption{Point cloud data sets used for performance benchmarking}
    \label{tab:data_sets}
\end{table}

\begin{table}
    \centering
    \begin{tabular}{|ccccccc|}
        \hline 
         & Threshold & Max. memory & {Memory ratio} & User time  & Min. user time & Ripser user time \\
         random16 & - & 6.73 (0.49) MB & 15.0 (1.01) & 2.87 (.07) s & 2.66s & 2.20s \\
         sphere3 & - & 3.80 (0.27) MB & 18.39 (1.33) & 0.22 (0.04) s & 0.16s & 0.40s \\
         fractal-r & - & 4.79 (0.26) MB & 1.53 (.08) & 3.80 (0.71) s & 3.04s & 2.96s \\
         dragon & - & 18.15 (1.72) MB & 18.78 (1.50) & 123.85 (22.60) s & 101.56s & 100.47s \\
         \hline 
         o3-1024 & 1.80 & 7.32 (0.47) MB & 1.90 (0.11) & 1.57 (0.24) s & 1.25s & 1.32s \\
         o3-4096 & 1.40 & 57.77 (7.42) MB & 2.63 (0.39) & 27.71 (3.99) s & 23.34s & 23.89s \\
         torus4 & 0.15 & 137.44 (5.78) MB & 1.28 (0.05) & 29.31 (2.90) s & 26.09s & 34.96s \\
                & 0.30 & 516.58 (6.67) MB & 1.66 (0.02) & 302.51  (47.06) s & 255.51s & 447.06s \\
        \hline 
                 &      &             & Ripser max. memory &  &  & \\
            torus4    & 2.82 & 57.61 GB & 326.93 GB$^*$ & 292 days, 6.52 hrs & - & 2729s$^*$ \\
            cylco     & -    & 269.79 MB & 134.72 GB & 1 hr, 18.87 min & - & 1 hr, 18.52 min \\
         \hline
    \end{tabular}
    \caption{Benchmarking results for outputs with low dependence on number of threads. The maximum memory ratio is given by $\frac{\text{ripser maximum resident size}}{\qaup \text{ maximum resident size}}$. Standard deviations for average quantities are reported in parentheses. The experiments reported in the final two rows were only run once with 63 threads. Ripser was unable to complete reduction of the larger torus4 example due to memory constraints, so we report final readouts before it threw an error marked by $^*$ for that case}
    \label{tab:benchmark_results}
\end{table}
\end{landscape}

\section{Machine learning performance} \label{sec:results}
Having established some theoretical and experimental stability properties of unreduced PDs, as well as information about the cost to compute them in practice, what remains is to investigate their usefulness for ML applications compared to FR diagrams. This section reports on several ML experiments conducted towards this end. The tasks were chosen to probe a range of data types, filtrations, and task difficulties. A code repository for reproducing the results reported here is available at \url{https://parkeredw.com/files/unred_pd_paper_files.zip}.

\subsection{Methods and Data}
In addition to software written to compute unreduced PDs, our experiments relied in various places on existing libraries for TDA and ML~\cite{bauer2017phat,gudhi:CubicalComplex,scikit-learn,gudhi:urm,scikittda2019}. For some experiments, all unreduced PDs of a particular type or homology degree were empty, i.e., all points were ephemeral. In these cases, the affected PD types or homology degrees were excluded from model training. 

\Cref{tab:experiment_table} details information about component variations tested for each ML task specified \Cref{ssec:shape_description,ssec:fmnist_class_description,ssec:brain_artery_description}. One experiment for these tasks was conducted for each combination of component variations.

\begin{figure}
    \centering
    \includegraphics[width=0.9\textwidth]{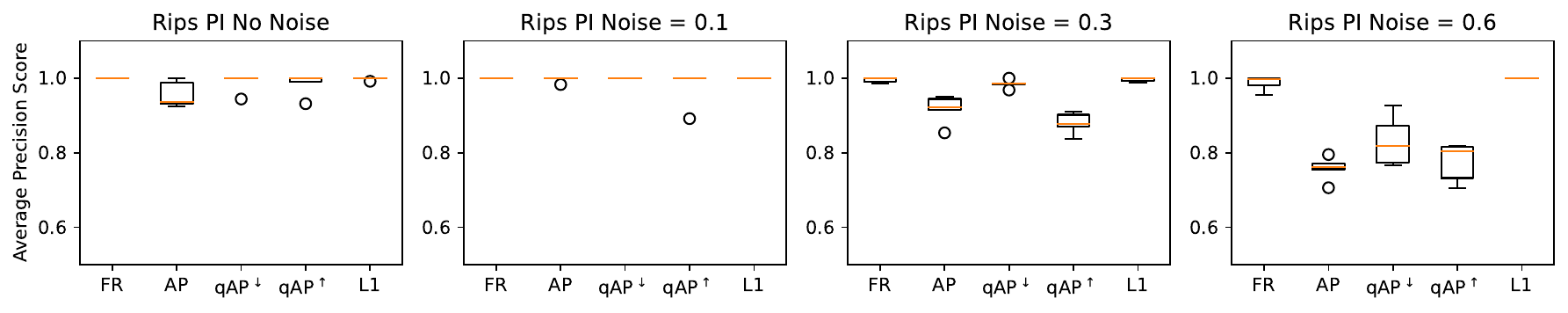}
    \caption{Average precision scores of random forest classifiers from synthetic shape classification experiments }
    \label{fig:shapeclassificationrips}
\end{figure}

\begin{table}
    \begin{center}
        \begin{tabular}{|c|c|c|c|c|}
            \hline
            \textbf{Data} & \textbf{Filtration} & \textbf{Types} &
            \textbf{Vectorizations} & \textbf{ML model} \\
            \hline
            Shape & alpha, Rips & L1, $\qadown$, ${\qaup}^*$, AP, FR
            & \multirow{3}{*}{\shortstack{$\bullet$ Persistence images\\  $\bullet$ Adcock Carlsson \\ coordinates}}
            & \multirow{3}{*}{\shortstack{Random\\ forest~\cite{Breiman2001}}} \\
            \cline{1-3}
            f-MNIST & ``Sweep'' & L1, FR & & \\
            \cline{1-3}
            Brain trees & See text & L1, $\qadown$, $\qaup$, AP, FR & & \\
            \hline
        \end{tabular}
    \end{center}
    \caption{Experiment component variations. The $\qaup$ diagrams for the Rips variant of the shape experiment used the slightly modified $\qaup$ construction described in \Cref{sec:compute} }
    \label{tab:experiment_table}
\end{table}

\subsubsection{Synthetic shape classification}\label{ssec:shape_description} To assess the capacity of unreduced diagrams to discriminate topologically distinct shapes, we generated synthetic data samples drawn from five subspaces of $\mathbb{R}^3$: a circle, two distant clusters, uniform random points, a sphere, and a torus. The ML task was to predict the shape label given a data sample as input. We compared the performance of random forest classifiers trained on vectorizations of unreduced diagrams to those trained on fully reduced diagrams, which are known to excel at this task. 

Each entry in this data set consisted of a point cloud of 200 points randomly sampled from one of the five shapes. These shapes were as similar in geometric scale as possible. Each of 50 point clouds generated per shape was labeled by shape. To increase the difficulty of the task, and assess model sensitivity to increasing noise, we also conducted variant experiments where every point in each point cloud was perturbed in a uniform random direction, drawn from the 2-sphere, with a uniform random magnitude drawn from (0,$\mu$) for each value $\mu=0.1, 0.3,$ and $0.6$. \Cref{fig:shapeclassificationrips} reports results for Rips filtrations and persistence images vectorizations and \Cref{fig:full_shapeclassification} reports results for both filtration types (alpha and Rips) and vectorization types (Adcock-Carlsson coordinates and persistence images).

\subsubsection{Fashion-MNIST image classification} \label{ssec:fmnist_class_description} The Fashion-MNIST dataset~\cite{xiao2017fashionmnist} consists of 70,000 grayscale images with $28\times 28$ pixel resolution. The images depict clothing items from 10 clothing classes. To create complexes from the images, we followed a similar cubical-complex protocol to one previously used by Adcock and Carlsson~\cite{adcock-carlsson} and Barnes et al.~\cite{perea2021}. Specifically, each image was filtered by ``sweeping'' separately in each of the four cardinal directions: each nonzero pixel was assigned a normalized value in $[0,1]$ that grew linearly from left-to-right, right-to-left, bottom-to-top, or top-to-bottom, depending on the direction of the sweep. Pixels which originally were 0 retained that value. From each sweep, filtered cubical complexes were constructed using the lower-star filtration~\cite{gudhi:CubicalComplex}. These complexes each consisted of 3249 total cubes: $784 = 28 \times 28$ 2-cubes corresponding to original image's pixels (whose filtration values were determined by the values assigned by a sweep), $1624 = 2\times 29\times28$ edges (their filtration values being determined by the smallest filtration value of any pixel each bounded), and $841=29\times 29$ 0-cubes (vertices, which inherited the smallest filtration value of their 4, or 2, incident edges). Each sweep produced a cube-wise filtration from which $H_0$ and $H_1$ PH classes were computed. Thus, a total of 8 PDs were generated for each image. \Cref{fig:fmnist_FI_and_APS} reports results.

\subsubsection{Brain artery tree regression}\label{ssec:brain_artery_description}
To compare reduced and unreduced diagrams on a difficult regression task, we reanalyzed a data set prepared by Bendich et al.~\cite{bendich2016} from 3D Magnetic Resonance Angiography brain images collected by Bullit et al.~\cite{bullitt2005}. Each of the 98 data entries estimated a subject's brain artery tree and consists of a 3D point cloud together with edges between points determined by a tube-tracking algorithm. To create a filtered complex, we largely followed the procedure of Bendich et al.~\cite{bendich2016}. In degree 0, vertices were filtered by height and edges had the same filtration value as the higher of their two vertices. In degree 1, we used alpha complexes constructed from Euclidean distances between vertices (i.e., ignoring edge adjacency). Each brain was also labeled with the subject's age. The supervised learning task was predicting a subject's age using their brain artery tree as input. \Cref{tab:brain_results} reports results.

\subsubsection{Hyperparameter tuning}\label{ssec:hyperparameters}

Hyperparameter tuning was conducted using 5-fold cross validation with \texttt{hyperopt}~\cite{bergstra2013hyperopt}. Only vectorization hyperparameters were tuned while random forest hyperparameters were not. Final performance assessment also used 5-fold cross validation with the best identified hyperparameters for every experiment. A hyperparameter tuning holdout set was used for the shape and Fashion-MNIST tasks, but this was not done for the brain artery regression as there were too few data entries available. 

\subsubsection{Feature importance for f-MNIST classification}\label{ssec:fmnist_feature_importance_description}
We also conducted a feature importance analysis with the f-MNIST dataset to further investigate what differences captured by FR and L1 PDs contribute to differences in testing performance for this data set. First, we simplified the experiment to make the FR and L1 models easier to compare. 

Using initial testing performances, we identified that the 2-class pair which most improved performance using L1 PDs as opposed to FR PDs was class 5 (sandals) vs class 7 (sneakers). We then trained L1 and FR random forest classification models as in the original experiment with persistence image vectorization and an 80\%/20\% training/testing split of the whole f-MNIST data set. The PI vectorization hyperparameters used were the same for both the L1 and FR models and were manually chosen based on simple properties of the filtration (e.g., all persistence pairs occupied the $[0,1]^2$ region of the birth-persistence plane). 

As a further regularization, the original 800 dimensional vectorizations were mean-centered, scaled by standard deviation in the standard way, and projected onto the first 13 principal components (PCs) determined from the training data before classification. The first 13 PCs were chosen as the minimum number which accounted for over 90\% of the variance in the FR vectorizations of the training data. 

Note that training does not change any model parameters except for centering and scaling until after the vectorization step. The top portion of \Cref{fig:fmnist_FI_and_APS}(c) and \Cref{fig:DifferenceBetweenClasses} record differences between average PIs for sandals and sneakers after centering and scaling but before PCA projection. 

We then computed permutation-based feature importance using all testing data, not just class 5 and class 7, for both models. This metric is computed feature-by-feature. For each feature, the values of the feature can be shuffled among all data samples, keeping the other features unshuffled. Whole model testing performance on this shuffled data is then computed, with the difference between unpermuted classification accuracy and permuted classification accuracy subsequently reflecting feature importance. A large positive feature importance in feature $k$ indicates that shuffling feature $k$ in the testing data among data samples resulted in  degraded performance relative to the original unshuffled testing data.  Repeating this procedure and averaging the results across many random permutations and all features yields a final measurement of feature importance for each feature. Abbreviated results are reported in~\Cref{fig:fmnist_FI_and_APS}(c) with full results in the Appendix (\Cref{fig:full_feature_importance}).

For each model, we inspected the single most important feature according to this analysis. The corresponding principal component is a vector of weights in the original 800-dimensional vector space. That space is determined by concatenation of the PIs across 4 sweeps and 2 homology dimensions. We reshaped the weights of these principal components to depict the relative contribution of PI regions to each PC's projected feature. \Cref{fig:reconstructed_weights} compares the most important principal components in the FR and L1 models.

\subsubsection{Additional technical details}

In most experiments, multiple PDs were computed for each data entry. PDs for homology of different degrees were counted as separate for this purpose. We used a standard method to vectorize, computing a single vector for each of a data entry's PDs, then concatenating the resulting vectors to obtain a final output. 

This procedure introduces a set of vectorization hyperparameters for each PD associated to a data entry. Our experiments allow each PD's vectorization hyperparameters to vary independently. For example, each synthetic shape point cloud had associated $H_0$, $H_1$, and $H_2$ PDs. Correspondingly, vectorization hyperparameters for all 3 homology degrees were allowed to vary independently.

Some hyperparameter choices for both persistence image and Adcock-Carlsson vectorization entail raising birth, death, or persistence coordinates of PD points to a positive power as part of computations. We observed that this can cause overflows of the 32-bit floats preferred by the Python library $\texttt{sklearn}$ when moderately large coordinates occur in the input PDs. This arose primarily for L1 diagrams computed from alpha complexes. Other filtration types cannot have especially large filtration values relative to inputs by construction. To mitigate the issue, we preprocessed outputs after vectorizing by replacing overflows with the maximum possible non-infinite representable 32-bit float in Python for the synthetic shape and f-MNIST experiments.

\subsection{Results \& Discussion}

We conducted two types of PH-ML experiments to assess the relative performance of our four unreduced PD constructions and so report on those results separately here. \Cref{ssec:results_class_and_regression} reports on the majority of the experiments, which focused on supervised ML performance when varying components in the PH-ML pipeline, particularly the type of PD used. Results for our feature importance experiment which investigated what differences captured by FR and unreduced PDs contribute to differences in downstream testing performance for a particular data set are discussed in \Cref{ssec:results_feature_importance}.

\subsubsection{Classification and regression experiments}\label{ssec:results_class_and_regression}
The ML tasks used for performance comparison experiments exhibited a range of difficulties. Across all shape classification experiments, even with relatively large $\mu = 0.6$ point cloud perturbations, some diagram type achieved at or near the maximum of 1.0 average precision score (\Cref{fig:full_shapeclassification}, rightmost column). The top performing L1 diagrams achieved a somewhat lower median average precision score of $0.86$ \Cref{fig:fmnist_FI_and_APS}) in our f-MNIST experiment. The highest average $R^2$ performance across brain artery experiments was a relatively low $0.3541$ with 1.0 being the maximum possible (\Cref{tab:brain_results}). 

In all experiments, at least one type of unreduced PD performed as well or better than fully reduced diagrams, according to the chosen metrics of model performance (\Cref{fig:shapeclassificationrips,fig:fmnist_FI_and_APS,fig:full_shapeclassification} and \Cref{tab:brain_results}). For example, L1-based regression models achieved an $R^2$ score of 0.3541 using persistence image vectorization while FR diagrams averaged 0.3445, although both exhibited high standard deviation of more than 50\% and 30\% of the average performance value across cross-validation splits respectively~(\Cref{tab:brain_results}). 

No PD type outperformed all others across all tasks. L1 or FR diagrams did, however, attain or tie for best performance across all of our experiments. In fact, across all filtration and vectorization types and noise levels, synthetic shape classification models trained on L1 PDs had a median average precision score within 0.6 percentage points of the FR models (\Cref{fig:full_shapeclassification}). The most pronounced performance difference between FR and L1 diagrams was in our f-MNIST classification experiment, where L1 diagrams outperformed FR diagrams by at least 10 percentage points using both ACC and PI diagram vectorizations (\Cref{fig:fmnist_FI_and_APS}(a)). 

AP diagrams were also able to achieve high synthetic shape classification performance at all noise levels using alpha filtrations, though they showed significant performance degradation using Rips complexes above moderate noise levels ($\mu \geq 0.3$) (\Cref{fig:full_shapeclassification}). Similarly, the $R^2$ scores of AP, $\qaup$, and $\qadown$ models showed notable overlap in their testing performance distributions across folds with the distributions for L1 and FR models, with a marginal reduction (approximately 0.03) in average $R^2$ score compared to the models trained on FR diagrams. 
 
Relative performance of different PD types remained largely similar while varying other pipeline components. This suggests that PD type was a primary contributor to performance differences. Although alpha complexes performed better than Rips complexes in our synthetic shape experiments (\Cref{fig:full_shapeclassification}), PD type performance rankings remained largely the same when varying filtrations. Albeit, some ties arose using alpha complexes that were not present with Rips complexes.

\begin{table}
    \centering
    \begin{tabular}{|c|c|c|}
        \hline
        \textbf{Vectorization} & \textbf{PD type} & \textbf{Average $R^2$} \\
        \hline       
        Persistence image  & \text{L1} & 0.3541 (0.19) \\
        & AP & 0.3196 (0.17) \\
         & $\qadown$ & 0.3130 (0.17)\\
         & $\qaup$ & 0.3165 (0.11) \\
         & FR & 0.3445 (0.12) \\
         \hline        
        Adcock-Carlsson & \text{L1} & 0.1844 (0.08) \\
        & AP & 0.2303 (0.16) \\
         & $\qadown$ & 0.2733 (0.13) \\
         & $\qaup$ & 0.2850 (0.10) \\
         & FR & 0.3042 (0.11) \\
         \hline
    \end{tabular}
    \caption{$R^2$ testing performance of regression pipeline for brain artery experiments averaged over 5 folds. Standard deviations are in parentheses }
    \label{tab:brain_results}
\end{table}

\subsubsection{Feature importance experiment}\label{ssec:results_feature_importance}

Analyzing what differences in PD types contribute to downstream performance differences is more challenging. Our simplified models used on f-MNIST data for feature importance comparison replicated the qualitative results from the earlier cross-validated experiment. Testing performance of the simplified models was 0.79 average precision score for the FR model and 0.91 for the L1 model. Restricted to class 5 and class 7 testing examples, testing performance was 0.90 for the FR model and 0.98 for the L1 model. 

\begin{figure}[t]
    \centering
    \includegraphics[width=0.75\linewidth]{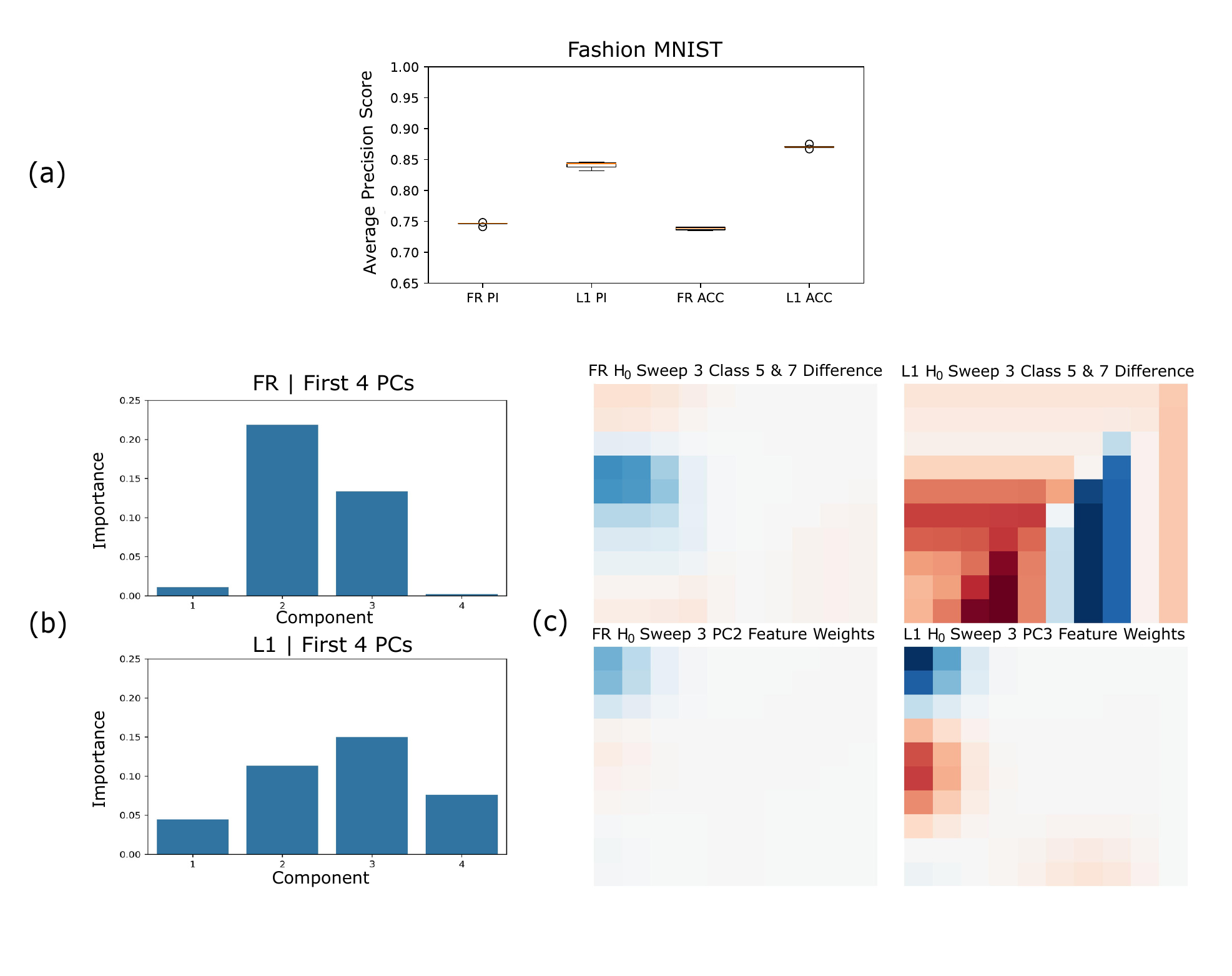}
    \caption{ (a) Average precision scores of random forest classifiers that were trained either on PI or AC coordinates of FR or L1 PDs derived from cubical filtrations of f-MNIST images. (b) Permutation feature importance of the first 4 principal components (PCs) of simplified FR (top) and L1 (bottom) f-MNIST classification models. (c) Top: Differences between average H$_0$-sweep 3 PIs for class 5 (sandals) and 7 (sneakers). Bottom: Reshaped weight vectors corresponding to FR principal component 2 (PC2) and L1 principal component 3 (PC3) for trained H$_0$-sweep 3 random forests  }
    \label{fig:fmnist_FI_and_APS}
\end{figure}

\begin{figure*}
    \centering
    \includegraphics[width=0.99\textwidth]{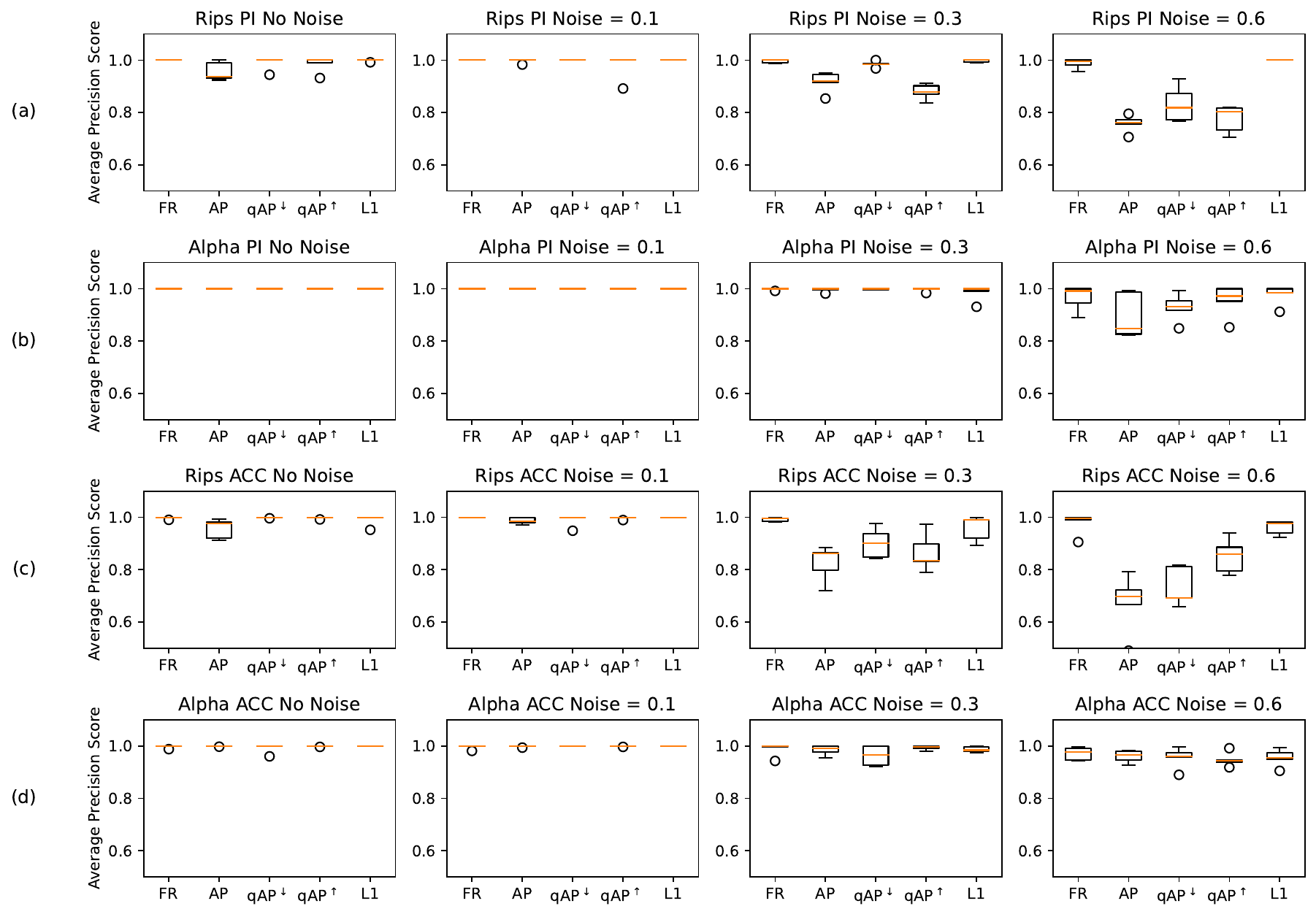}
    \caption{Average precision scores of random forest classifiers that were trained on (a) PIs derived from Rips filtrations, (b) PIs derived from alpha filtrations, (c) AC coordinates derived from Rips filtrations, and (d) AC coordinates derived from alpha filtrations. Filtrations were built on point cloud samples (from 5 shape classes), that were perturbed with varying amounts of noise 
    }

    \vspace*{3in}
    
    \label{fig:full_shapeclassification}
\end{figure*}

\Cref{fig:fmnist_FI_and_APS,fig:DifferenceBetweenClasses} from our f-MNIST feature importance experiment visually indicate that the average difference between class 5 and class 7 PIs was greater for L1 diagrams than FR (\Cref{fig:DifferenceBetweenClasses}). A permutation test for whether the class 5 and 7 average PIs were distinct using 10000 permutations yielded a $p$-value of $10^{-4}$ for both the FR examples and L1 examples. Moreover, the $L_2$ norm of the difference between the average scaled PIs of these classes was only 3.49 in the FR case but was 14.18 in the L1 case.

\begin{figure*}
    \centering
    \includegraphics[width=0.8\textwidth]{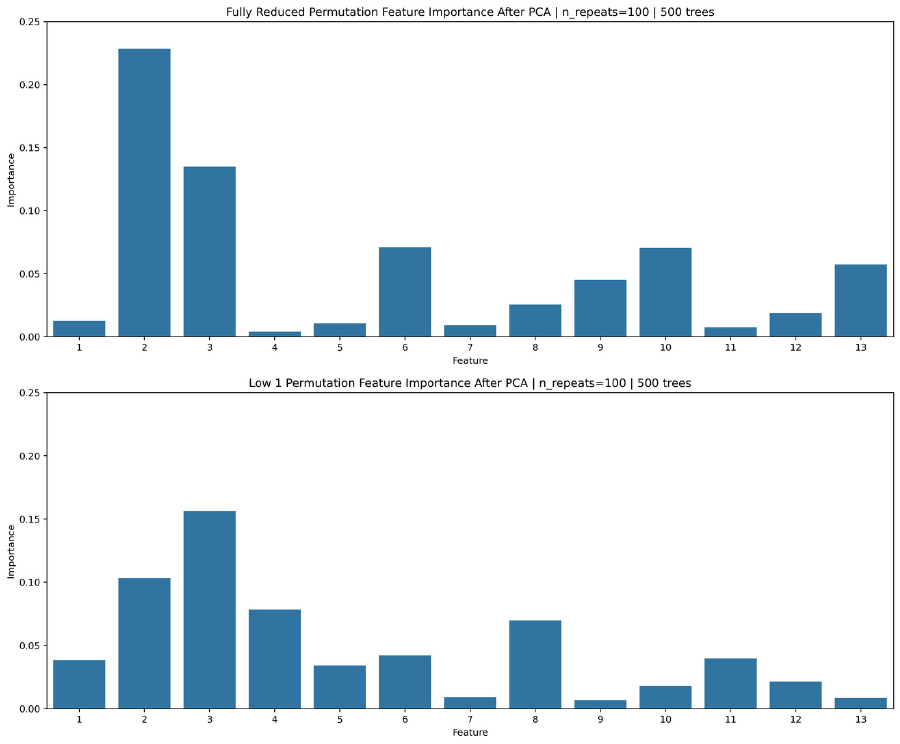}
    \caption{Permutation feature importance per component for the FR (top) and L1 (bottom) models. The importance score of a component increases proportionately to the average amount of performance degradation after randomly permuting only the corresponding component among all samples \label{fig:full_feature_importance}}

    \vspace*{3in}
    
\end{figure*}

\begin{figure*}
    \centering
    \includegraphics[width=0.99\textwidth]{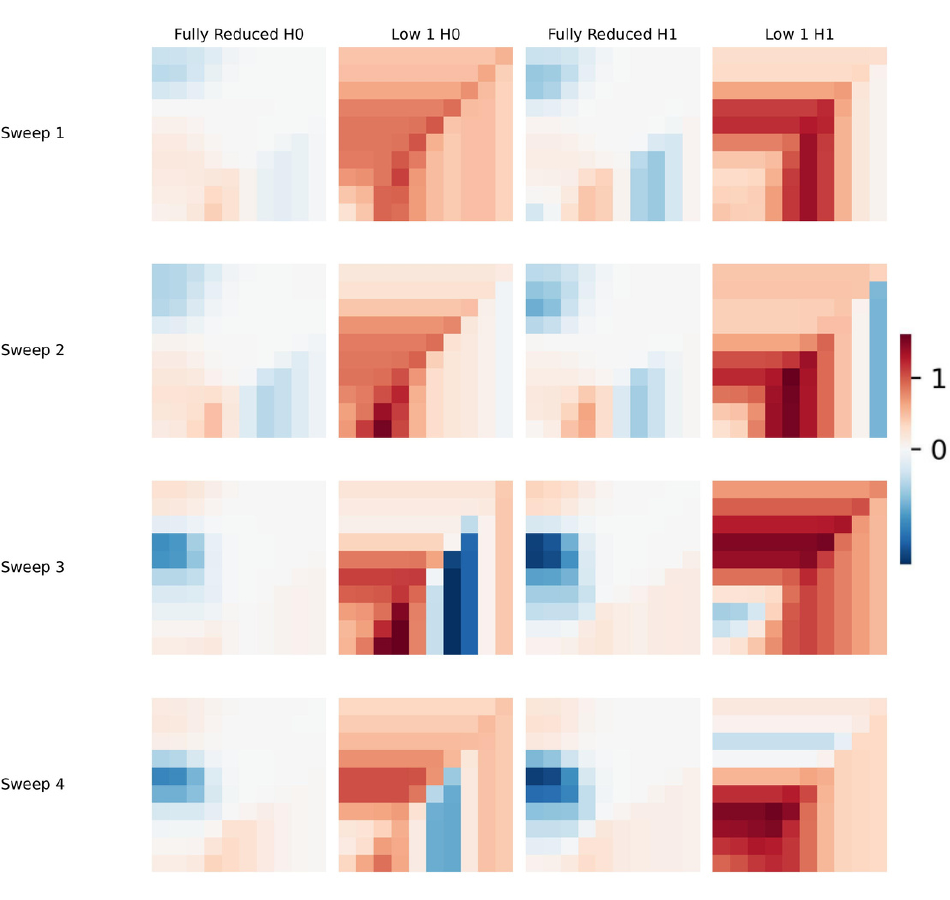}
    \caption{The differences between the average class 7 PIs and average class 5 PIs of corresponding sweep numbers and homology dimensions. Averages were taken over all PIs in the testing set after normalization via mean centering and standard deviation scaling  \label{fig:DifferenceBetweenClasses}}

    \vspace*{3in}

\end{figure*}

Feature importance results indicate that the features in L1 and FR vectorizations that are most useful useful for classification occur in distinct regions of the corresponding PIs. The most important principal component features were prominent in both the FR and L1 case (PC2 and PC3 respectively), exhibiting substantially higher importance than the second most important feature (\Cref{fig:full_feature_importance}). The cosine similarity between the corresponding most important FR and L1 principal components was $0.35$, indicating that these PCs focused on different regions of persistence image space. 

\begin{figure*}
    \centering
    \includegraphics[width=0.99\textwidth]{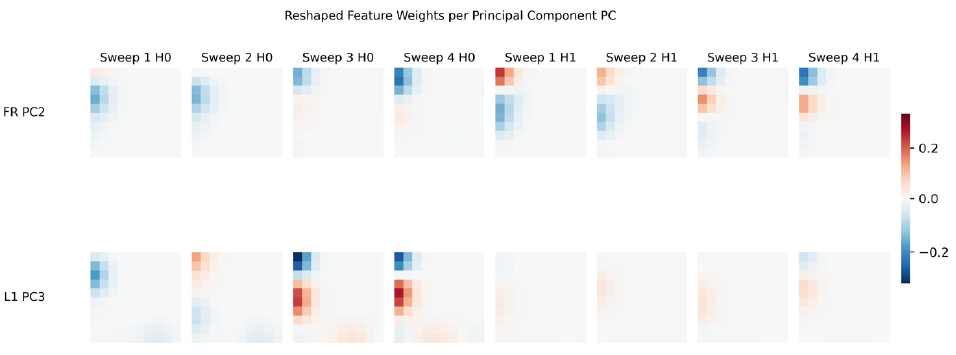}
    \caption{Reconstructed feature weight vectors corresponding to most important principal components with the largest positive permutation feature importance score per model type. The second principal component (PC2) for the FR PD and the third principal component (PC3) for the L1 PD were most important to the held-out testing data prediction score. The color map scale is normalized by the minimum and maximum weight values across all feature weights and model types to allow comparison of the scales of the weights of FR PC2 and L1 PC3 \label{fig:reconstructed_weights}}

    \vspace*{3in}
    
\end{figure*}

Interestingly, the relative importance of features derived from different homological dimensions also appears to change with diagram type. For instance, there appear to be significant weights corresponding to features in each of the 8 component FR PDs (covering both $H_0$ and $H_1$), while in the L1 diagrams, the most important principal component has coefficients mostly concentrated in $H_0$ diagrams (\Cref{fig:reconstructed_weights}). 

\section{Conclusion}

An interesting, and somewhat surprising, conclusion of this study is that L1 diagrams performed as well or better than FR diagrams in all cases. That said, both QA and L1 PDs are somewhat less interpretable than the FR and AP diagrams. As discussed in \Cref{sec:background}, the AP PDs do not capture any information in most homology degrees for popular flag-type filtrations (e.g., Vietoris-Rips). Thus, the promising experimental results reported here motivate further exploration of other unreduced PD constructions, specifically related to the quasi-apparent pair diagrams, which avoid this limitation.

Another tantalizing observation of this study is that the instability of AP diagrams may correspond to extremely rare data configurations. Almost without exception, stability results in the TDA literature are global Lipschitz statements which require universal control of metric expansion. Our stability experiments (\Cref{fig:Stability_Experiment_Plots}) motivate adopting alternative measures of instability which more accurately capture the expected behavior of transformations into topological representations. 

The range and number of experiments represents a significant limitation of this study. While the experiments we performed covered several different data types and exhibited a variety of difficulty levels, they still represent a relatively small number of contexts compared to the overall corpus of existing PH-ML applications. The relative performance of diagram types may be problem specific and additionally depend on factors not evaluated in this work (e.g., classifier/regressor architecture).  

A major goal of this study was to elucidate the computational benefits of avoiding the (full) reduction of simplex-wise filtrations to generate true PDs. Using optimized implementations and parallelization, we observed significant memory and compute time savings over state-of-the art methods for performing reduction on Rips filtrations. That said, we observed significantly longer times to vectorize L1 diagrams than to vectorize other diagram types (not reported). This was due to the larger number of persistence pairs in these diagrams. Careful implementation and proper benchmarking will be required to properly characterize the computational trade-offs due to vectorizing larger diagrams while avoiding diagram reduction and is the focus of ongoing research. 

Indeed, it is \emph{a priori} necessary to store each non-ephemeral pair in memory while computing an unreduced PD, both as a step in \Cref{alg:qap} and to output the resulting unreduced diagram. This could be a large number of pairs. Note, however, that most PD vectorization methods only require aggregation, not storage, of results from small computations on each point in a PD. Designing algorithms which only store vectorizations of unreduced diagrams or approximations of them is a natural direction for future study.

Altogether, these results suggest that unreduced PDs can serve as effective substitutes for fully reduced PDs in PH-ML pipelines, at least in terms of task performance. It also appears that, in some tasks, unreduced diagrams can encode usefully different information than fully reduced diagrams. The impact of incorporating features from both unreduced and fully reduced diagrams is beyond the scope of this paper and is the subject of ongoing research.

\bibliographystyle{abbrv}
\bibliography{refs}

\begin{figure*}
    \centering
    \includegraphics[width=0.75\linewidth]{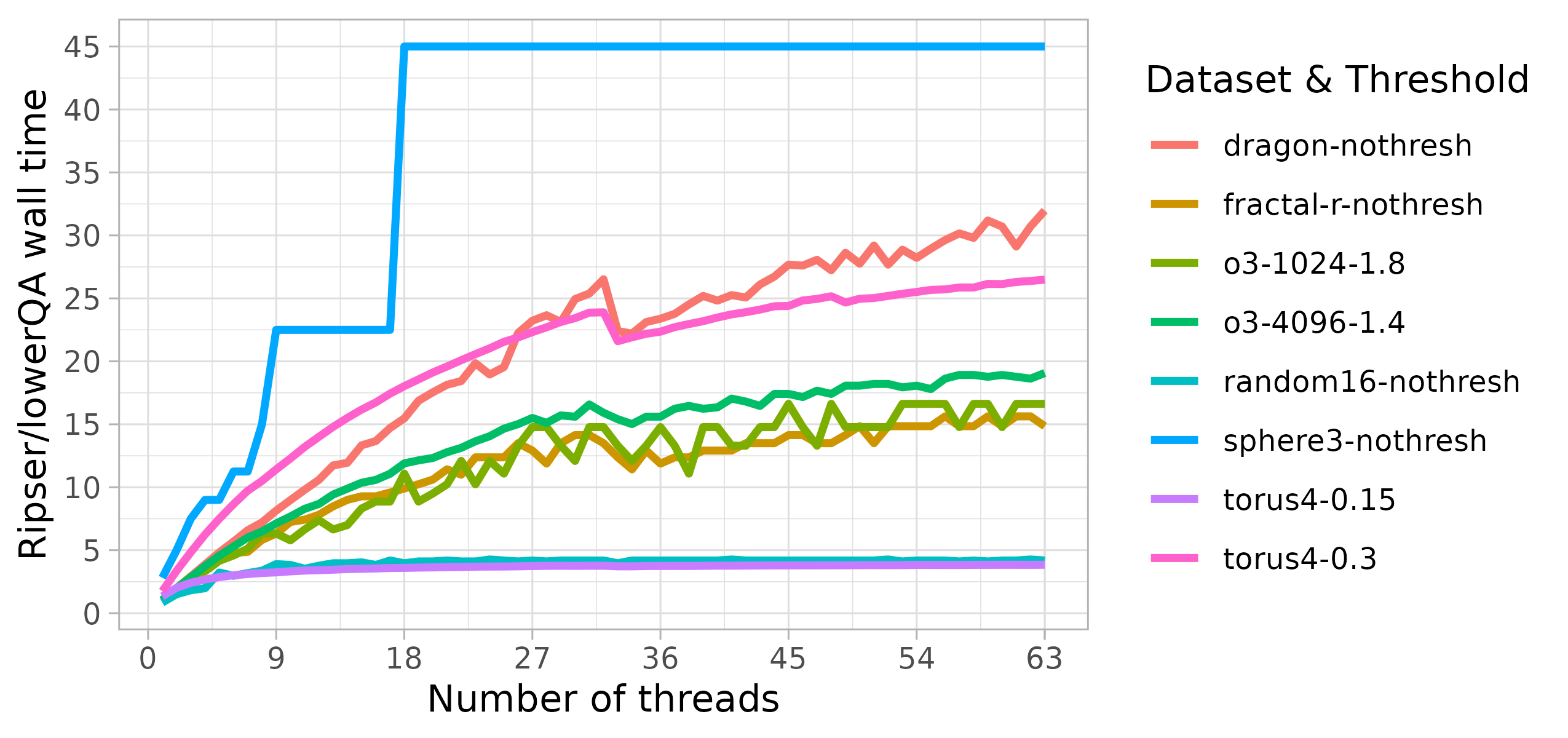}
    \caption{\Cref{fig:walltime_ratios} without smoothing}
    \label{fig:unsmoothed_wall_time}
\end{figure*}

\end{document}